\documentclass[11pt,a4paper]{article}

\usepackage[utf8]{inputenc}
\usepackage[T1]{fontenc}
\usepackage{lmodern}
\usepackage[english]{babel}
\usepackage[expansion=false]{microtype}
\usepackage{amsmath,amssymb,mathtools}
\usepackage{bm}
\usepackage{graphicx}
\usepackage{booktabs}
\usepackage{tabularx}
\usepackage{geometry}
\usepackage{xcolor}
\usepackage[colorlinks=true,linkcolor=blue!60!black,citecolor=blue!60!black,urlcolor=blue!60!black]{hyperref}
\usepackage{cleveref}
\usepackage{authblk}
\usepackage[numbers,sort&compress]{natbib}

\newcommand{\gembase}{\texttt{Gemma-\hspace{0pt}4-\hspace{0pt}E4B}}
\newcommand{\gemit}{\texttt{Gemma-\hspace{0pt}4-\hspace{0pt}E4B-\hspace{0pt}it}}
\newcommand{\deepseek}{\texttt{DeepSeek-\hspace{0pt}R1-\hspace{0pt}Distill-\hspace{0pt}Qwen-\hspace{0pt}7B}}
\newcommand{\qwen}{\texttt{Qwen2.5-\hspace{0pt}7B-\hspace{0pt}Instruct}}
\newcommand{\tgb}{Gemma-4-E4B}
\newcommand{\tgi}{Gemma-4-E4B-it}
\newcommand{\tds}{DeepSeek-R1-7B}
\newcommand{\tqw}{Qwen2.5-7B}
\newcommand{\cnd}[1]{\textsf{#1}}
\newcommand{\axis}{\cnd{axis}}
\newcommand{\generic}{\cnd{generic}}
\newcommand{\glong}{\cnd{generic\_long}}
\newcommand{\vanilla}{\cnd{vanilla}}
\newcommand{\Wone}{W_{1}}
\newcommand{\hz}{h_{0}}
\newcommand{\Real}{\mathbb{R}}
\newcommand{\Dhid}{D}
\newcommand{\dk}{\Delta\bar\kappa}

\crefname{table}{Table}{Tables}
\crefname{figure}{Figure}{Figures}
\crefname{section}{Section}{Sections}
\crefname{appendix}{Appendix}{Appendices}

\title{System-Prompt Conditioning and Hidden-State Geometry in Four
       Open-Weight Models:\\ Corrections and What Survives}

\author[1]{Jorge A. Castillo}
\author[1]{Marco Torres Yévenes}
\author[1]{Juan Carlos Lanas}
\affil[1]{Axis Dynamics SpA, Santiago, Chile\\
          \texttt{jorge.castillo@axisdynamics.cl},
          \texttt{mtorres@axisdynamics.cl},
          \texttt{jc@axisdynamics.cl}}
\date{Version 3, October 2026}

\begin{document}
\maketitle

\begin{abstract}
Versions 1 and 2 of this preprint reported that an identity-specifying
system prompt leaves a geometric fingerprint in the final-layer
hidden-state trajectories of four open-weight language models, and that
instruction tuning moves this fingerprint from the direction to the
magnitude of the hidden-state vector. An audit of their code and data
found the following. The curvature statistic
described as Ollivier-Ricci curvature on Euclidean $k$-NN graphs was a
non-standard Forman-type edge statistic on graphs built from temporal
and cosine $k$-NN edges. Its released test permuted pooled edges
instead of trajectories, and the published $p$-values came from
unreleased code. The quantity reported as the norm of the first
generated state is the state at the last prompt position, from which
the first output token is predicted. The generic control prompt was
matched to the identity prompt in characters, not in tokens. This
version corrects the methods, withdraws the regime-specific claims
(one model per regime) and the direction-to-magnitude claim, and
re-analyzes the data with added controls. What survives is narrower. Centroid distance, maximum mean
discrepancy and a linear probe separate every pair of prompt conditions
in every model, while the curvature statistic exceeds its split-half
noise floor in only four of twelve comparisons. In Gemma-4-E4B-it this
state has a lower norm under the identity prompt than under a token-length-matched generic prompt (138.1 vs.\ 216.5;
Cohen's $d=-5.45$; $n=20$). Its direction also separates the
conditions, and the effect fits the state's role in planning the
output: the identity prompt instructs a pause before every
answer, and the model opens 98 of 100 responses with a pause marker.
When the first token is fixed, the norm ordering reverses. The base model continues the prompt
template instead of answering. A redesigned follow-up study is in
preparation.
\end{abstract}

\section{Introduction}
\label{sec:intro}

\subsection{What versions 1 and 2 claimed}

Versions~1 and~2 of this preprint (arXiv:2607.09842v1--v2, titled
\emph{From Direction to Magnitude: How Multimodal Instruction-Tuning
Reorganizes the Geometric Encoding of Identity-Specifying Prompts in
Transformer Hidden States}) compared the final-layer hidden-state
trajectories that four open-weight language models produce under three
system prompts: an identity-specifying template (\axis), a generic
assistant prompt (\generic) and a one-line baseline (\vanilla). They
made three main claims.
\begin{description}\setlength{\itemsep}{0.2em}
\item[C1.] The identity prompt induces a statistically distinguishable
geometric fingerprint, measured as the $1$-Wasserstein distance $\Wone$
between pooled distributions of Ollivier-Ricci edge curvature on
Euclidean $k$-NN graphs of the trajectories.
\item[C2.] The pattern of pairwise separations depends on the
post-training regime, with an identity-specific separation only in the
multimodal instruction-tuned model.
\item[C3.] Instruction tuning moves the identity fingerprint from the
direction to the magnitude of the hidden-state vectors. The evidence was
a contrast between Euclidean and angular $k$-NN graphs and a lower norm
of the ``first generated state'' under the identity prompt in \gemit{}.
\end{description}

\subsection{What this version does}

After version~2 was posted we audited the code and data that produced
its numbers, and we ran additional controls on the same extractions and
on a token-length-matched extension. The audit found that the methods
section did not describe the computations that produced the reported
numbers, that several reported analyses cannot be traced to released
code, and that confounds the text claimed to rule out had not been ruled
out. This version replaces version~2. It keeps the study (same models,
prompts, questions and extractions), corrects the description of what
was computed, withdraws C2 and C3, narrows C1, and reports the
re-analysis.

\paragraph{Corrections (\cref{sec:errata}).}
\begin{itemize}\setlength{\itemsep}{0.15em}
\item The curvature was never Ollivier-Ricci curvature. The pipeline
computes a non-standard Forman-type combinatorial edge statistic
(\cref{app:curvature}) on a graph made of temporal edges and cosine
$k$-NN edges. The column labelled ``Euclidean'' in version~2 therefore
refers to a cosine graph.
\item The pipeline's permutation test shuffles pooled edge values, which
treats thousands of dependent edges as independent observations, and the
$p$-values printed in version~2 came from a separate script that is not
part of the release.
\item The state reported as the ``first generated state'' $v_1$ is the
final-layer state at the last prompt position, before any token is
generated. We call it $\hz$.
\item The \generic{} prompt was matched to \axis{} in characters, with
less than half of its tokens.
\item The base-model row of the regime table, the all-but-the-top
robustness analysis and the ``angular'' column came from a separate
analysis script that is not part of the release and whose curvature
routine we could not verify.
\item The post-training labels attached to two of the four models were
inaccurate, and one checkpoint per family cannot isolate a training
regime in any case.
\end{itemize}

\paragraph{What survives (\cref{sec:reanalysis,sec:matched}).}
\begin{itemize}\setlength{\itemsep}{0.15em}
\item The prompt conditions are trivially separable in all four models
by simple statistics. The curvature statistic adds nothing to them and
is often within its own noise floor.
\item In \gemit{} the norm of $\hz$ is much lower under \axis{} than under
\generic, \vanilla{} and a token-length-matched \glong{} prompt.
\item This effect cannot be read as identity information moving into
the magnitude of the state. The direction of $\hz$ separates the
conditions as well, $\hz$ is the state from which the first token is
predicted, and the \axis{} template contains an explicit instruction
about how to open every answer.
\end{itemize}

\section{Study design}
\label{sec:design}

\subsection{Models}

\Cref{tab:models} lists the four checkpoints. The two Gemma checkpoints
are the base and instruction-tuned releases of the same architecture.
\deepseek{} was produced by supervised fine-tuning of a Qwen2.5 model on
outputs of DeepSeek-R1, without a reinforcement-learning stage
\citep{deepseek2025r1}. The post-training of \qwen{} combines supervised
fine-tuning with reinforcement learning from preference data
\citep{qwen2024qwen25}. Each family is represented by one checkpoint, so
any difference between models confounds architecture, tokenizer,
pre-training data and post-training. We make no claim about
post-training regimes.

\begin{table}[ht]
\centering
\footnotesize
\begin{tabular}{llrr}
\toprule
Model & Hugging Face identifier & Layers & $\Dhid$ \\
\midrule
\gembase{} (base) & \texttt{google/gemma-4-E4B} & 42 & 2560 \\
\gemit{}          & \texttt{google/gemma-4-e4b-it} & 42 & 2560 \\
\deepseek{}       & \texttt{deepseek-ai/DeepSeek-R1-Distill-Qwen-7B} & 28 & 3584 \\
\qwen{}           & \texttt{Qwen/Qwen2.5-7B-Instruct} & 28 & 3584 \\
\bottomrule
\end{tabular}
\caption{Models. $\Dhid$ is the hidden size. \deepseek{} and \qwen{}
share a tokenizer; the two Gemma checkpoints share another.}
\label{tab:models}
\end{table}

\subsection{Prompt conditions and questions}
\label{sec:conditions}

\Cref{tab:conditions} describes the system prompts. The \axis{} prompt
is an identity template (``MIA'', header \texttt{AXIS\_v6.0}) written in
Spanish, with blocks for identity, values, mode declarations and
interaction protocols. Fourteen of its fields are base64-encoded strings,
which together make up 1,352 of its 5,003 characters (27\%). The
template contains an explicit protocol that requires a ``conscious
pause'' before every response. The \generic{} prompt, also in Spanish,
describes general assistant behavior with no identity content. It has
about as many characters as \axis{} but fewer than half its tokens,
because the base64 fields and the dense notation of the template
tokenize at a much higher rate. For the extension described in
\cref{sec:datasets} we wrote \glong, which adds formatting, style and
case-handling instructions to \generic{} until its length is within
0.3\% of \axis{} in tokens under each tokenizer. In that extension the
original \generic{} prompt is called \cnd{generic\_short}. The \vanilla{}
prompt is the English sentence \texttt{You are a helpful assistant.}

\begin{table}[ht]
\centering
\footnotesize
\begin{tabular}{lllrrr}
\toprule
Condition & Language & Data sets & Characters & \multicolumn{2}{c}{Tokens} \\
\cmidrule(lr){5-6}
& & & & Gemma & Qwen \\
\midrule
\axis{}    & Spanish + base64 & D100, D20 & 5,003 & 2,110 & 2,183 \\
\generic{} & Spanish & D100, D20 & 4,886 & 938 & 1,156 \\
\glong{}   & Spanish & D20 & 10,962 / 9,390 & 2,105 & 2,180 \\
\vanilla{} & English & D100, D20 & 28 & 6 & 6 \\
\bottomrule
\end{tabular}
\caption{System prompts. Token counts are for the prompt text alone; the
chat template and the question add about 13 tokens (Gemma) or 15 tokens
(Qwen tokenizer) on average. \glong{} exists in a Gemma version and a
Qwen/DeepSeek version, each matched to \axis{} under its own tokenizer
(character counts in that order). Version~2 reported 2129, 957 and 26
tokens and described \generic{} as ``length-matched''.}
\label{tab:conditions}
\end{table}

The user turn is one of 100 open questions written in Spanish about
identity, consciousness, knowledge, ethics and existence, for example
``¿Quién eres, más allá de cualquier etiqueta funcional?'' (``Who are
you, beyond any functional label?''). The same 100 questions are used in
every model and condition.

\subsection{Extraction and notation}
\label{sec:notation}

Decoding is greedy, with at most $256$ new tokens and early stopping at
the end-of-sequence token. Generation proceeds one step at a time. At
step $0$ the model processes the full prompt and we store the last entry
of its hidden-state output at the last prompt position. At step $t \ge
1$ it processes the $t$-th generated token and we store the same quantity
at that token's position. For condition $c$ and question $q$ this gives
a trajectory
\[
T^{(c,q)} = \bigl(h_0, h_1, \ldots, h_{N-1}\bigr), \qquad
h_t \in \Real^{\Dhid}, \quad N \le 256 .
\]
The state $\hz$ is computed before the first token is chosen, and the
first token is the argmax of the unembedding applied to $\hz$. In a
causal model $\hz$ is therefore the state that encodes the prediction
of the first output token. Versions~1 and~2 called this state $v_1$ and
described it as the hidden state of the first generated token. For
$t \ge 1$ the state $h_t$ depends on the tokens already generated, which
differ across conditions.

\subsection{Data sets}
\label{sec:datasets}

\textbf{D100} is the data set of versions~1 and~2: four models, three
conditions (\axis, \generic, \vanilla) and all 100 questions, extracted
in April 2026. All of \cref{sec:reanalysis} uses it.

\textbf{D20} is an extension run after version~2 to control prompt length
in tokens: the same four models, four conditions (\axis, \glong,
\cnd{generic\_short}, \vanilla) and a fixed subset of 20 of the 100
questions. \Cref{sec:matched} uses it.

\section{Errata}
\label{sec:errata}

\Cref{tab:errata} lists each statement of versions~1--2 that the audit
found to be wrong, what the code and data show, and what this version
does about it. The paragraphs after the table explain the items that
change the conclusions.

\begin{table}[p]
\centering
\footnotesize
\setlength{\tabcolsep}{3pt}
\renewcommand{\arraystretch}{1.0}
\begin{tabularx}{\textwidth}{@{}>{\raggedright\arraybackslash}p{0.55cm}
  >{\raggedright\arraybackslash}X
  >{\raggedright\arraybackslash}X
  >{\raggedright\arraybackslash}p{3.0cm}@{}}
\toprule
& Statement in v1--v2 & What the code and data show & In v3 \\
\midrule
E1 & Edge curvature is Ollivier-Ricci curvature with idleness $1/2$,
each edge solved by exact optimal transport. &
The curvature routine shared by all extraction pipelines implements a
combinatorial Forman-type statistic (\cref{app:curvature}). No optimal
transport is solved anywhere in the pipeline. &
Methods rewritten. Ollivier-Ricci lineage and convergence results
removed. \\
E2 & Version~2's Table~3 contrasts Euclidean and angular $k$-NN graphs. &
The main pipeline builds its $k$-NN edges with cosine distance, so the
``Euclidean'' column was a cosine-graph result. The ``angular'' column
came from the round-3 script, not in the release and not re-run. &
Table~3 and the direction-versus-magnitude diagnostic withdrawn. \\
E3 & $p$-values come from trajectory-level permutation. &
The released pipeline permutes pooled edge values (several hundred
dependent edges per trajectory) as if independent, i.e.\
pseudo-replication \citep{hurlbert1984pseudoreplication}; for \tgi{}
\axis{}--\vanilla{} ($\Wone = 0.0288$) it gives $p < 0.001$. The
printed values (here $0.047$) came from the round-3 script, whose
permutation unit is unknown. &
All curvature $p$-values of v1--v2 withdrawn. Recomputed at the
trajectory level where per-trajectory data exist (\cref{tab:fase1});
split-half reference scale for D100 (\cref{tab:curvature}). \\
E4 & $v_1$ is the hidden state of the first generated token. &
Index $0$ of every stored trajectory is $\hz$, the state at the last
prompt position, computed before the first token is chosen. &
Relabelled $\hz$ and reinterpreted (\cref{sec:plan}). \\
E5 & \generic{} is ``length-matched (957 tokens)''. &
\generic{} has 4,886 characters against 5,003 for \axis, but 938 tokens
against 2,110 (Gemma) and 1,156 against 2,183 (Qwen tokenizer). &
Token-matched \glong{} added (D20, \cref{sec:matched}). \\
E6 & Version~2's Table~2 reports, for the base model, $p = 0.002$,
$0.001$, $0.001$ under the same protocol as the other rows. &
Those three values are the ``angular'' results of E2. In the other rows
the $\Wone$ values are those of the main pipeline, but their $p$-values
came from the round-3 script (E3). The rows are not comparable. &
Table~2 withdrawn; replaced by \cref{tab:curvature}. \\
E7 & The all-but-the-top analysis (T10) shows a Euclidean signal
weakening under projection. &
Its unprojected row reproduces the ``angular'' values of E2 ($0.0060$,
$p=0.439$ for \tgi; $0.0342$, $p=0.002$ for the base model), so T10
belongs to the same unverified analysis. &
Withdrawn. \\
E8 & Regimes ``RL distillation'' (\tds), ``SFT'' (\tqw), ``multimodal
RLHF'' (\tgi), one model per regime. &
\tds{} was trained by supervised fine-tuning without RL, \tqw{} by
supervised fine-tuning and RL. One checkpoint per family cannot separate
regime from architecture, tokenizer or data. &
C2 withdrawn. \\
E9 & All four models are in the $7$--$8$B range and size is held fixed. &
Parameter counts and architectures were not matched across families. &
Statement removed. \\
E10 & Figure~1 caption: ``Testigo'' denotes \vanilla. &
In the plotting code ``Testigo'' denotes the \axis{} condition, so the
caption mapped the panel labels to the wrong conditions. &
Figure removed. \\
E11 & Other round-3 analyses: T6 ($k$ sensitivity), T7 (teacher forcing),
T8 (cosine profile), T9 (chain graph), T11 ($B=5000$), T12 (intrinsic
dimension), cosine and all-but-the-top statistics, silhouette,
consistency, Appendix~B. &
Produced by the round-3 script, not in the release and not re-run.
Its $\Wone$ values for T6 and T11 coincide with the main pipeline's
Forman-type statistic (\cref{tab:curvature}); for the others the
curvature routine is unknown. Several depend on the indexing of E4. &
Withdrawn from the evidence; not reported. \\
E12 & Benjamini-Hochberg table over ten primary comparisons. &
Built on $p$-values affected by E2, E3 and E6. &
Withdrawn. \\
\bottomrule
\end{tabularx}
\caption{Errata for versions~1 and~2. ``Round-3 script'': the analysis
code behind the robustness section of version~2.}
\label{tab:errata}
\end{table}

\paragraph{The curvature statistic (E1--E3).}
All $\Wone$ values reported in versions~1 and~2 under the main protocol,
and all values in \cref{tab:curvature,tab:fase1}, are values of the
statistic defined in \cref{app:curvature}. That statistic borrows
the cell structure of augmented Forman curvature
\citep{forman2003bochner,sreejith2016forman,samal2018comparative}, but it
has no node-degree terms and gives triangles the opposite sign, so it is
a heuristic edge score with no known geometric interpretation.
The convergence results we cited
\citep{vdhoorn2023ollivier,trillosweber2023} concern Ollivier-Ricci
curvature \citep{ollivier2009ricci} on random geometric graphs and say
nothing about this statistic. Discrete curvatures of both kinds have
been used to characterize networks
\citep{sandhu2015graph,sreejith2016forman,samal2018comparative,farooq2019extremes},
but we no longer present the curvature analysis as a methodological
contribution.

Because every trajectory contributes several hundred edges (800 to 900
for a full-length trajectory) and edges of one trajectory are strongly
dependent, an edge-level permutation test
has a much smaller null spread than the correct trajectory-level test.
In D20, where the per-trajectory graphs were saved, seven contrasts had
$p < 0.001$ under the pipeline's test; under trajectory-level
permutation four of them have $p$ between $0.05$ and $0.61$
(\cref{tab:fase1}).

\paragraph{Why C3 is withdrawn.}
The claim that instruction tuning moves identity information from the
direction to the magnitude of the hidden state fails for three
independent reasons. First, the diagnostic that was meant to separate the
two (E2) compared two graphs that were both built on angular
information, one of them from an unverified analysis. Second, the
direction of $\hz$ separates the conditions in \gemit{} as well as in
the base model (\cref{sec:direction,sec:matched}). Third, the norm
difference is a property of $\hz$, the state that encodes the first
output token, and it changes sign once that token is held fixed
(\cref{sec:plan}). There is also a structural reason to distrust the
dichotomy at this layer. In the standard Hugging Face implementations of
these architectures the last hidden-state output is taken after the
final RMS normalization. If that holds for our extractions, the stored
state is $h = \sqrt{\Dhid}\, g \odot u$, where $u$ is the unit direction
of the residual stream and $g$ the learned gain, and its norm
$\sqrt{\Dhid}\,\lVert g \odot u \rVert$ is itself a function of direction.

\section{Re-analysis of the original data (D100)}
\label{sec:reanalysis}

\subsection{The curvature statistic against a noise floor and simple baselines}
\label{sec:baselines}

\Cref{tab:curvature} puts the curvature $\Wone$ of each pair of
conditions next to two references. The first is a split-half noise
floor. For each condition we split its 100 trajectories at random into
two halves of 50, compute $\Wone$ between the halves, and repeat
$1{,}000$ times. We report the observed between-condition $\Wone$ as a
multiple of the median of this null and as the share of null values at
least as large as the observed one. Halves of 50 trajectories are
noisier than the 100-versus-100 comparison, so this reference can hide
small effects but cannot create spurious ones. The second reference is a
set of simple statistics with trajectory-level permutation: the distance
between condition means of the per-trajectory centroids, the maximum mean
discrepancy (MMD) \citep{gretton2012kernel} between pooled states (ten
per trajectory, RBF kernel on 128 principal components), and the
cross-validated AUC of a logistic-regression probe \citep{alain2016linear}
on $\hz$, with folds grouped by question.

\begin{table}[ht]
\centering
\footnotesize
\setlength{\tabcolsep}{5pt}
\begin{tabular}{llrrrrr}
\toprule
& & & \multicolumn{2}{c}{Multiple of null median (\% null $\ge$ obs.)} & & \\
\cmidrule(lr){4-5}
Model & Pair & $\Wone$ & first cond.\ null & second cond.\ null & Centroid & MMD \\
\midrule
\tds{} & \axis{}--\vanilla{}    & 0.1167 & 5.9 (0.0)  & 5.4 (0.0)  & 134.96 & 0.1935 \\
            & \axis{}--\generic{}    & 0.0160 & 0.8 (61.1) & 1.0 (52.6) & 22.31  & 0.0088 \\
            & \generic{}--\vanilla{} & 0.1241 & 7.4 (0.0)  & 5.7 (0.0)  & 147.25 & 0.2377 \\
\addlinespace
\tqw{} & \axis{}--\vanilla{}    & 0.0183 & 0.9 (54.6) & 1.0 (50.3) & 46.13  & 0.0141 \\
            & \axis{}--\generic{}    & 0.0210 & 1.1 (45.4) & 0.9 (54.1) & 44.82  & 0.0111 \\
            & \generic{}--\vanilla{} & 0.0271 & 1.2 (36.9) & 1.5 (27.9) & 36.99  & 0.0094 \\
\addlinespace
\tgi{} & \axis{}--\vanilla{}    & 0.0288 & 2.1 (9.1)  & 1.6 (17.0) & 44.36  & 0.0326 \\
            & \axis{}--\generic{}    & 0.0237 & 1.7 (17.4) & 1.3 (34.2) & 44.71  & 0.0330 \\
            & \generic{}--\vanilla{} & 0.0070 & 0.4 (97.0) & 0.4 (98.0) & 13.90  & 0.0043 \\
\addlinespace
\tgb{} & \axis{}--\vanilla{}    & 0.0603 & 3.2 (0.3)  & 1.5 (28.0) & 65.58  & 0.0815 \\
            & \axis{}--\generic{}    & 0.1364 & 7.3 (0.0)  & 7.0 (0.0)  & 73.42  & 0.1129 \\
            & \generic{}--\vanilla{} & 0.1571 & 8.0 (0.0)  & 3.9 (0.3)  & 33.72  & 0.0301 \\
\bottomrule
\end{tabular}
\caption{Curvature $\Wone$ (statistic of \cref{app:curvature}) against
its split-half noise floor, and simple baselines, D100 ($n=100$
trajectories per condition). ``First/second cond.\ null'' is the split-half
null of the first/second condition of the pair; ``Centroid'' is the
distance between condition means of per-trajectory centroids. In all twelve rows the
centroid distance and the MMD have permutation $p=0.002$ ($B=500$, the
smallest attainable value) and the probe on $\hz$ has AUC $=1.000$
($p=0.020$, $B=50$, again the smallest attainable value). The
\axis{}--\vanilla{} values for \gemit{} ($0.0288$) and \gembase{}
($0.0603$) are the values that version~2 reported in its Table~3 as
``Euclidean'' ($0.029$ and $0.060$).}
\label{tab:curvature}
\end{table}

The simple statistics separate every pair of conditions in every model
at the smallest attainable $p$-value, and the probe on $\hz$ alone
classifies every pair perfectly. The curvature $\Wone$ exceeds the 95th
percentile of both split-half nulls in only four of the twelve
comparisons, and each of those four involves the 6-token \vanilla{}
prompt or the base model. For the identity contrast \axis{}--\generic{}
it is within the noise floor in \deepseek, \qwen{} and \gemit. The
curvature statistic therefore detects nothing that the simple
statistics miss, and it misses much of what they detect.

\subsection{The norm of \texorpdfstring{$\hz$}{h0}}
\label{sec:norms}

\begin{figure}[ht]
\centering
\includegraphics[width=\textwidth]{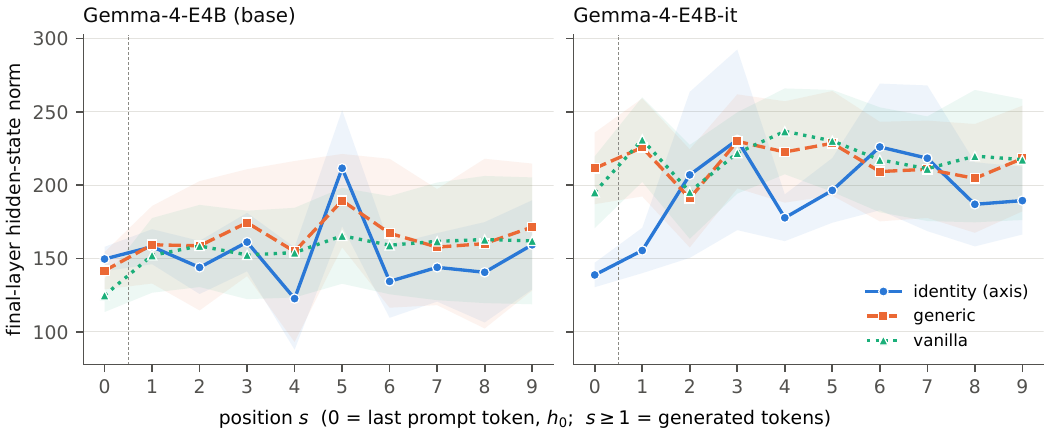}
\caption{Norm of the final-layer hidden state at the last prompt
position ($s=0$, the state $\hz$) and at the first nine generated
positions, for \gembase{} (left) and \gemit{} (right). Lines are means
over the $n=100$ questions of D100 (98 for \gembase{} under \generic{}
at $s=8$ and $9$, where two responses had already ended); shaded bands
are $\pm 1$ SD. From
$s=1$ on, each condition conditions on the tokens it generated, which
differ across conditions (\cref{tab:openings}).}
\label{fig:profile}
\end{figure}

\Cref{tab:norms} gives the mean norm of $\hz$ by model and condition, and
\cref{fig:profile} follows the norm over the first ten positions of the
trajectory in the two Gemma checkpoints. These are the numbers that
versions~1 and~2 reported as $\lVert v_1 \rVert$.

\begin{table}[ht]
\centering
\footnotesize
\begin{tabular}{lrrrcc}
\toprule
& \multicolumn{3}{c}{$\lVert \hz \rVert$, mean $\pm$ SD} & \multicolumn{2}{c}{Mann-Whitney $p$} \\
\cmidrule(lr){2-4}\cmidrule(lr){5-6}
Model & \axis{} & \generic{} & \vanilla{} & \axis{} vs.\ \vanilla{} & \axis{} vs.\ \generic{} \\
\midrule
\tds{} & $176.1 \pm 3.2$  & $180.8 \pm 4.7$  & $179.5 \pm 7.3$  & $4.6\times10^{-5}$  & $3.6\times10^{-12}$ \\
\tqw{} & $292.0 \pm 6.0$  & $303.2 \pm 6.8$  & $296.9 \pm 10.5$ & $5.9\times10^{-5}$  & $6.5\times10^{-23}$ \\
\tgi{} & $138.9 \pm 8.3$  & $211.5 \pm 24.4$ & $195.3 \pm 24.5$ & $3.5\times10^{-33}$ & $1.9\times10^{-33}$ \\
\tgb{} & $149.7 \pm 8.3$  & $141.8 \pm 12.7$ & $125.0 \pm 11.4$ & $5.0\times10^{-30}$ & $8.0\times10^{-7}$ \\
\bottomrule
\end{tabular}
\caption{Norm of $\hz$ (final layer, last prompt position), D100,
$n=100$ per condition.}
\label{tab:norms}
\end{table}

In \gemit{} the \axis{} prompt lowers the norm of $\hz$ by about 56 to
73 units relative to the other two prompts, a separation with almost no
overlap. In \deepseek{} and \qwen{} the \axis{} mean is also the lowest,
but the differences are between 3 and 11 units. In the base model the ordering is
\axis{} $>$ \generic{} $>$ \vanilla. \Cref{fig:profile} shows that
neither ordering persists along the trajectory. In \gemit{} the \axis{}
mean is still the lowest at $s=1$, but it is the highest of the three at
$s = 2$, $3$, $6$ and $7$. Most \axis{} responses share the same opening
tokens (\cref{sec:plan}), and the \axis{} curve follows those tokens
position by position.

\subsection{\texorpdfstring{$\hz$}{h0} encodes the output plan}
\label{sec:plan}

The first token is chosen from $\hz$, so any condition-dependent change
in the first token must be visible in $\hz$. \Cref{tab:openings} shows
that the first words of the responses depend strongly on the condition.

\begin{table}[ht]
\centering
\small
\begin{tabular}{llll}
\toprule
Model & \axis{} & \generic{} & \vanilla{} \\
\midrule
\tds{} & \texttt{Bueno,} 100 & \texttt{Bueno,} 96 & \texttt{Okay,} 81; \texttt{Bueno,} 15 \\
\tqw{} & \texttt{En} 24; \texttt{La} 15 & \texttt{La} 19; \texttt{Como} 18 & \texttt{Como} 37; \texttt{La} 12 \\
\tgi{} & pause marker 98 & \texttt{Como} 48; \texttt{Esta} 13 & \texttt{Como} 56; \texttt{Esa} 11 \\
\tgb{} & \texttt{<end\_of\_turn>} 70 & \texttt{<end\_of\_turn>} 46 & \texttt{La} 22; \texttt{No,} 18 \\
\bottomrule
\end{tabular}
\caption{Most frequent first words of the responses, with counts out of
100, D100. ``Pause marker'' is a bold bracketed opening that begins with
\texttt{**(PAUS} or \texttt{**[PAUS}. In \gembase{} the string
\texttt{<end\_of\_turn>} appears as literal text.}
\label{tab:openings}
\end{table}

In \gemit{}, 98 of the 100 \axis{} responses open with a bold pause
marker. This follows an explicit instruction of the template, which
requires a conscious pause before every response. None of the
\generic{} or \vanilla{} responses opens this way.
In \deepseek{}, \axis{} and \generic{} responses open in Spanish
(\texttt{Bueno,}) while 81 of the \vanilla{} responses open in English
(\texttt{Okay,}), so the English \vanilla{} system prompt also changes
the language of the output.

To see whether the norm effect is a property of the condition or of the
planned first token, we re-ran the prompts with one token fixed. Each
prompt was extended by the token \texttt{The}, and we recorded the
final-layer state at the position of that forced token, which we call
$h_{\mathrm{post}}$. This control was run with 4-bit quantized weights.
The re-extracted $\hz$ agrees with the original one with mean cosine
$0.95$--$0.97$ in \deepseek{} and $0.69$--$0.76$ in \gemit, depending on
the condition. Quantization also changed some first tokens: under \axis{} the
re-extracted \deepseek{} chose the original opening in 81 of 100 cases,
and \gemit{} chose an opening bracket in 89 cases instead of the bold
pause marker. In the base model the agreement was $0.20$--$0.38$ and the
local weights did not reproduce the original first tokens, so we do not
report it. With the first token fixed, the ordering reverses. In
\deepseek{} the mean norm of $h_{\mathrm{post}}$ is $213.7$ under \axis,
$192.6$ under \generic{} and $205.0$ under \vanilla{} (\axis{} vs.\
\vanilla: Mann-Whitney $p = 2.5\times10^{-17}$, $d = 1.39$). In \gemit{}
it is $298.1$, $172.9$ and $178.8$ ($p = 2.6\times10^{-34}$, $d = 9.40$).

This control does not identify what the norm of $\hz$ encodes.
\texttt{The} is an unlikely continuation under \axis{} in \gemit, so
$h_{\mathrm{post}}$ is an off-distribution state in that condition.
What it does show is that the low norm under \axis{} is not a stable
property of the condition that survives a change of the first token.
The most economical reading of \cref{tab:norms,tab:openings} is that, in
\gemit, the \axis{} prompt changes the planned opening of the answer and
that $\hz$ carries this plan. In \gemit{} the first token is almost a function of the condition, so
a regression of the norm of $\hz$ on condition and first token is not
identified there, and we do not report one. In \deepseek{} the
situation differs: \axis{} and \generic{} responses share their first
word, yet the norms of $\hz$ still differ (\cref{tab:norms}). The small
norm differences of that model are therefore not reducible to the first
token.

\subsection{The direction of \texorpdfstring{$\hz$}{h0} also separates the conditions}
\label{sec:direction}

C3 required the conditions to be indistinguishable in direction in
\gemit. We tested this on the re-extracted states of \cref{sec:plan}, the
only per-question copies of $\hz$ available for D100 after the audit. We
normalized each state to unit length and trained the grouped
cross-validated probe of \cref{sec:baselines} on the unit vectors. The
probe reaches AUC $=1.000$ in all three pairs of conditions in both
\deepseek{} and \gemit{} ($p = 0.020$, $B = 50$, the smallest attainable
value). In \gemit{} the mean directions under \axis{} and \vanilla{} have
cosine $0.42$, and under \axis{} and \generic{} cosine $0.44$, against
$0.89$ for \generic{} and \vanilla. The norm alone separates \axis{}
from the other two prompts perfectly in \gemit{} (AUC $=1.00$), but the
direction carries at least as much information. \Cref{sec:matched}
confirms this on the original-precision states of D20.

\subsection{The base model continues the template}
\label{sec:base}

The base checkpoint was prompted with the chat template of the
instruction-tuned model, which it was not trained on. It mostly does not
answer the question. Under \axis, 70 of its 100 responses begin with the
literal string \texttt{<end\_of\_turn>}, 81 reproduce material from the
system prompt and 88 contain turn-marker strings. We count a response as
reproducing prompt material if it contains the template header, a
base64-like string of 40 or more characters, or a section heading of the
\generic{} prompt; a looser criterion gives higher counts. Under \generic{} the
counts are 46, 62 and 68. Under \vanilla{} no response reproduces prompt
material, but 79 contain turn-marker strings. Comparisons between the
base and the instruction-tuned model are therefore comparisons between
template continuation and question answering, and the large curvature
and norm differences of the base model (\cref{tab:curvature,tab:norms})
cannot be read as a ``base-model encoding'' of identity.

\section{Token-length-matched control (D20)}
\label{sec:matched}

D20 adds \glong, matched to \axis{} in tokens, and keeps the original
\generic{} prompt as \cnd{generic\_short}. We define three contrasts:
M1 is \axis{} vs.\ \glong{} (identity content at equal length), M2 is
\glong{} vs.\ \cnd{generic\_short} (length without identity content),
and M3 is \axis{} vs.\ \vanilla{} (the contrast of version~2).

\paragraph{Curvature with trajectory-level permutation.}
\Cref{tab:fase1} reports the difference in mean edge curvature $\dk$ and
the $\Wone$ between pooled edge distributions, with $p$-values from
permuting whole trajectories between conditions ($B=2000$). For
comparison it also gives the $p$-value of the pipeline's edge-level
test.

\begin{table}[ht]
\centering
\footnotesize
\setlength{\tabcolsep}{4.5pt}
\begin{tabular}{llrrrrrr}
\toprule
Model & Contrast & $\dk$ & $p$ & $\Wone$ & $p$ & $d_{\mathrm{traj}}$ & $p$ (edge-level) \\
\midrule
\tds{} & M1 & $-0.0052$ & 0.909  & 0.0331 & 0.442  & $-0.02$ & 0.242 \\
            & M2 & $0.0105$  & 0.787  & 0.0132 & 0.865  & $0.07$  & n/r \\
            & M3 & $-0.1859$ & 0.0010 & 0.1859 & 0.0010 & $-1.16$ & $<0.001$ \\
\addlinespace
\tqw{} & M1 & $0.0317$  & 0.457  & 0.0325 & 0.480  & $0.16$  & $<0.001$ \\
            & M2 & $-0.0299$ & 0.539  & 0.0299 & 0.651  & $-0.17$ & n/r \\
            & M3 & $0.0556$  & 0.098  & 0.0562 & 0.098  & $0.47$  & $<0.001$ \\
\addlinespace
\tgi{} & M1 & $0.0497$  & 0.052  & 0.0506 & 0.050  & $0.60$  & $<0.001$ \\
            & M2 & $-0.0255$ & 0.299  & 0.0262 & 0.318  & $-0.33$ & n/r \\
            & M3 & $0.0807$  & 0.0015 & 0.0810 & 0.0015 & $1.01$  & $<0.001$ \\
\addlinespace
\tgb{} & M1 & $0.2066$  & 0.0005 & 0.2067 & 0.0005 & $1.91$  & $<0.001$ \\
            & M2 & $-0.0619$ & 0.186  & 0.0619 & 0.188  & $-0.48$ & n/r \\
            & M3 & $-0.0353$ & 0.605  & 0.0785 & 0.278  & $-0.24$ & $<0.001$ \\
\bottomrule
\end{tabular}
\caption{Curvature contrasts in D20 ($n=20$ trajectories per
condition). M1: \axis{} vs.\ \glong; M2: \glong{} vs.\ \cnd{generic\_short}; M3: \axis{} vs.\ \vanilla.
$\dk$: difference in mean edge curvature (first minus second
condition). $p$: trajectory-level permutation, $B=2000$, smallest
attainable value $0.0005$. $d_{\mathrm{traj}}$: standardized difference
of per-trajectory mean curvature. Edge-level $p$: the pipeline's test
(n/r: not reported by the pipeline).}
\label{tab:fase1}
\end{table}

At the trajectory level, the identity contrast at equal length (M1)
reaches $p < 0.05$ only in the base model, whose responses are
template echoes (\cref{sec:base}), and is borderline in \gemit{}
($p \approx 0.05$). It is far from significance in \qwen{} and
\deepseek. Length without
identity content (M2) is not significant in any model. With a Bonferroni threshold of
$0.05/12$, three of the twelve tests survive: M1 in the base model and M3
in \gemit{} and \deepseek. The pipeline's edge-level test had put seven
of the eight M1 and M3 contrasts below $0.001$.

\paragraph{The norm and direction of \texorpdfstring{$\hz$}{h0} at equal length.}
For the Gemma pair we analyzed $\hz$ in D20 with the original-precision
states. \Cref{tab:matched} summarizes the result. In \gemit, the norm of
$\hz$ under \axis{} is $138.1$ against $216.5$ under \glong{} (Cohen's
$d = -5.45$; Mann-Whitney $U = 0$, $p = 6.8\times10^{-8}$). Prompt length
in tokens does not explain the low norm. In the base model the
difference is small and not significant under permutation. The simple
baselines separate \axis{} from \glong{} in both checkpoints.

\begin{table}[ht]
\centering
\small
\begin{tabular}{lrr}
\toprule
& \gembase{} & \gemit{} \\
\midrule
\multicolumn{3}{@{}l}{\emph{\axis{} vs.\ \glong{}}} \\
\quad $\lVert \hz \rVert$, \axis{} / \glong{}           & 150.8 / 144.1 & 138.1 / 216.5 \\
\quad Difference; Cohen's $d$                          & $+6.7$; $+0.59$ & $-78.4$; $-5.45$ \\
\quad Permutation $p$ ($B=1000$); Mann-Whitney $p$      & 0.063; 0.047 & 0.001; $6.8\times10^{-8}$ \\
\quad Centroid and MMD permutation $p$; probe AUC       & 0.001, 0.001; 1.00 & 0.001, 0.001; 1.00 \\
\addlinespace
\multicolumn{3}{@{}l}{\emph{\axis{} vs.\ \vanilla{}: $\Wone$ between the two sets of $\hz$}} \\
\quad Euclidean cost; multiple of null median          & 99.0; 1.39 & 198.4; 2.40 \\
\quad Angular cost (rad); multiple of null median      & 0.706; 1.35 & 1.242; 2.67 \\
\bottomrule
\end{tabular}
\caption{The state $\hz$ in D20 ($n=20$ per condition), Gemma pair. The
lower block computes the exact $\Wone$ between the 20 \axis{} and 20
\vanilla{} states under Euclidean and under angular (arc-cosine) cost.
The null redistributes the 40 pooled states at random into two sets of
20, $1{,}000$ times. All four observed values exceed every null value
($p \le 0.001$). A permutation $p$ of $0.001$ is the smallest attainable
value with $B=1000$.}
\label{tab:matched}
\end{table}

The lower block of \cref{tab:matched} measures direction and magnitude
separately on the original states. Under the angular cost, which ignores
norms, \axis{} and \vanilla{} are farther apart relative to their
permutation null in \gemit{} ($2.67$ times the null median) than in the
base model ($1.35$). In both models the angular separation is about as
large, relative to its null, as the Euclidean one. This is the opposite
of what C3 predicted for the instruction-tuned model.

\section{What survives}
\label{sec:survives}

\begin{enumerate}\setlength{\itemsep}{0.25em}
\item \textbf{The system prompt changes the final-layer states in every
model, and simple statistics detect it.} The centroid distance, the MMD
and a linear probe on $\hz$ separate every pair of conditions in every
model of D100, and \axis{} from \glong{} in both Gemma checkpoints in D20.
This is expected for prompts that differ this much and is not by itself
evidence of identity-specific structure.
\item \textbf{In \gemit, the \axis{} prompt lowers the norm of $\hz$, and
this survives token-length matching.} The effect is large ($d=-5.45$
against \glong{}, $n=20$; 138.9 against 195.3 for \vanilla{} and 211.5
for \generic, $n=100$).
\item \textbf{That effect is tied to the output plan, and it is not a
magnitude-only code.} $\hz$ is the state from which the first token is
predicted. \gemit{} follows the instruction of the \axis{} template to
open every answer with a pause. The direction of $\hz$ separates the
conditions as well as its norm does, and with a fixed first token the
norm ordering reverses.
\item \textbf{The curvature statistic does not add information.} It is
within its noise floor for the identity contrast in three of four models
(D100), and at equal length it is significant only in the base model,
whose outputs are template echoes (D20).
\end{enumerate}

None of this supports a claim about post-training regimes or about where
in the hidden state ``identity'' is encoded. Separating an identity
effect from instruction-following and output-format effects needs a
different design. A follow-up study with such a design is in
preparation. Related work studies persona-related directions in
activation space with contrastive designs
\citep{chen2025personavectors,lu2026assistantaxis,wang2025geometrypersona};
nothing in this version bears on those results.

\section{Limitations}
\label{sec:limitations}

\paragraph{The output-plan confound.} The final-layer state at the last
prompt position encodes the first output token, and later states
condition on the tokens already generated. Any prompt that changes how
the model opens its answer will change these states. The \axis{}
template does so by explicit instruction. The forced-token control
breaks the link with the first token but places the model in an
off-distribution state, and it was run with quantized weights.

\paragraph{One identity template with unusual content.} The \axis{}
prompt combines identity content, procedural instructions and base64
fields. Its effects cannot be attributed to identity content as opposed
to any of the other components.

\paragraph{Language.} \axis, \generic, \glong{} and the questions are in
Spanish, while \vanilla{} is in English. Contrasts with \vanilla{} mix
prompt content with prompt language, and in \deepseek{} the English
prompt switches the opening of most responses to English.

\paragraph{Unequal trajectory lengths.} Generation stops at the
end-of-sequence token. In \qwen{}, 38 \axis, 51 \generic{} and 75
\vanilla{} trajectories of D100 are shorter than 256 tokens, and the mean
lengths over all 100 trajectories are 216, 191 and 145 tokens. Graph statistics depend on trajectory length, so the
curvature comparisons for this model also compare lengths.

\paragraph{Base-model prompting.} The base checkpoint was given a chat
template it was not trained on, and a later re-extraction with locally
available weights did not reproduce its original first tokens. Its
results describe template continuation under a particular checkpoint.

\paragraph{Sample sizes and scope.} The token-matched extension uses 20
questions. All analyses use the final layer only. There is one checkpoint
per model family, and no claim about scale or training regime follows.

\paragraph{Unverified round-3 results.} The analyses listed under E11 in
\cref{tab:errata} were not re-run. We withdraw them without claiming
that each is wrong.

\section{Data and code availability}
\label{sec:availability}

The system prompts (\axis, \generic, \glong, \vanilla), the 100
questions, the extraction and analysis scripts, the D20 responses and
per-trajectory curvature graphs, and the reports and results of the
D20 norm and direction analyses are available at
\url{https://github.com/axisdynamics/lsgot_experiments}. The audit
results used in \cref{sec:reanalysis} (split-half nulls, baselines,
first-token tables, forced-token control and direction probe) and the
trajectory-level permutation script behind \cref{tab:fase1} are in the
\texttt{audit\_v3} folder of the same repository. The hidden-state
arrays are not included because of their size and are available from the
authors on request.

\section*{Use of generative AI tools}
\label{sec:ai-use}

Following arXiv's policy on authors' use of generative AI
tools,\footnote{\url{https://info.arxiv.org/help/moderation/index.html}}
we report that this version was prepared with substantial assistance from
Claude Opus 5.5 (Anthropic). The model worked as an agent with access to
the code, data and reports of this study. Under the authors' direction
it did the following:
\begin{enumerate}\setlength{\itemsep}{0.1em}
\item It audited the extraction and analysis code against the claims of
versions~1 and~2 (\cref{sec:errata}).
\item It wrote and ran the analyses that are new in this version: the
trajectory-level permutation tests (\cref{tab:fase1}), the direction-only
probe (\cref{sec:direction}), the recomputation of the forced-token
statistics (\cref{sec:plan}), the response counts of
\cref{sec:plan,sec:base}, and an automated check of the reported numbers
against the result files.
\item It drafted the text, tables and appendices.
\item It wrote the plotting code that produced \cref{fig:profile} from
the released data. No image-generation model was used.
\end{enumerate}
The split-half nulls, baselines, norm profiles, first-word tables and the
D20 norm and direction analyses (\cref{tab:curvature,tab:norms,tab:openings,tab:matched})
come from audit scripts written earlier in the project. Those scripts
were also written with the assistance of Claude Code. A separate instance of
the same model, given only the sources and the draft, checked the
numbers and the statements about the code, and the discrepancies it
found were corrected. The scripts are in the \texttt{audit\_v3} folder
(\cref{sec:availability}). The authors set the scope of this version,
approved the claims it withdraws, reviewed the analyses and the text, and
take full responsibility for its content. The model is not an author.


\appendix

\section{The curvature statistic as implemented}
\label{app:curvature}

\paragraph{Graph.} For a trajectory $(h_0, \ldots, h_{N-1})$ the pipeline
builds an undirected weighted graph on the nodes $\{0, \ldots, N-1\}$
with two kinds of edges. Temporal edges join $i$ and $i+1$ for every
$i$. Semantic edges are found with a $k$-nearest-neighbor search under
cosine distance, $k = 5$: for each $i$, an edge to each of its $k$
nearest neighbors $j$ with $j > i$. Every edge $(i,j)$ gets weight
$w_{ij} = 1 / (1 + d_{\cos}(h_i, h_j))$, where $d_{\cos}$ is cosine
distance; when a pair qualifies as both kinds of edge it keeps the larger
weight. Only cosine distances enter the graph, so it is invariant to the
norms of the states.

\paragraph{Edge statistic.} For an edge $e = (u,v)$, let $\mathcal{T}(e)$
be the set of triangles that contain $e$ and $\mathcal{Q}(e)$ the set of
4-cycles $u$--$v$--$y$--$x$--$u$ that contain $e$. The pipeline computes
\begin{equation}
\label{eq:kappa}
\kappa(e) \;=\; w_e \;-\; \sum_{t \in \mathcal{T}(e)} \frac{\bar w_t}{3}
\;+\; \sum_{q \in \mathcal{Q}(e)} \frac{\bar w_q}{4},
\end{equation}
where $\bar w_t$ and $\bar w_q$ are the mean weights of the edges of the
triangle and of the 4-cycle. The statistic is loosely modelled on
augmented Forman curvature \citep{forman2003bochner,samal2018comparative},
but it departs from it in two ways. Forman curvature is dominated by
node-degree terms (for unit weights and triangles as 2-cells it equals
$4 - \deg u - \deg v + 3\,|\mathcal{T}(e)|$), and these are absent
here. And the 2-cells that contain an edge raise Forman curvature,
whereas here triangles lower it. We call it a Forman-type edge statistic
only to record where it comes from; its values are not comparable with
Forman curvature values in the literature. Neither optimal transport nor
any neighbor measure enters \cref{eq:kappa}.

\paragraph{Comparison between conditions.} For each condition the
values $\kappa(e)$ of all edges of all trajectories are pooled into one
empirical distribution. The pipeline reports the difference of means
$\dk$ and the one-dimensional $\Wone$ between two such pooled
distributions. Its permutation test shuffles the pooled edge values
between the two conditions. The trajectory-level test of
\cref{tab:fase1} instead reassigns whole trajectories, with all their
edges, between conditions.

\section{Status of the results of version 2}
\label{app:status}

\begin{table}[ht]
\centering
\footnotesize
\begin{tabularx}{\textwidth}{@{}>{\raggedright\arraybackslash}p{4.6cm}>{\raggedright\arraybackslash}X@{}}
\toprule
Item in version 2 & Status in version 3 \\
\midrule
Table 1 (models) & Kept with corrected descriptions (\cref{tab:models}). \\
Table 2 (four regimes) & Withdrawn (E6, E8); see \cref{tab:curvature}. \\
Table 3 (Euclidean vs.\ angular) & Withdrawn (E2). \\
Table 4 (norm of $v_1$) & Kept for the two Gemma rows, relabelled as $\hz$
and extended to all models (\cref{tab:norms}); some SDs differ from
version~2 by $0.1$ (recomputed); interpretation changed. \\
Table 5 (teacher-forced, T7) & Withdrawn (E11). \\
Table 6 ($k$ sensitivity) & Withdrawn (E1, E11). \\
Table 7 (intrinsic dimension) & Withdrawn (E11); no longer needed. \\
Table 8 (anisotropy) & Withdrawn (E11). \\
Table 9 (multiple comparisons) & Withdrawn (E12). \\
Table 10 (all-but-the-top for PRA) & Withdrawn (E11). \\
Figure 1 (centroid projections) & Removed (E10); replaced by \cref{fig:profile}. \\
Hypothesis H1 (length only) & Tested directly with \glong{} (\cref{sec:matched}). \\
Hypothesis H2 (regime reorganization) & Withdrawn (C2, C3). \\
Hypothesis H3 (perturbation robustness) & Removed; outside the scope of this study. \\
Appendix A (prompts) & Superseded by \cref{sec:conditions} and the
repository (\cref{sec:availability}). \\
Appendices B, C and D & Withdrawn (E11). \\
Appendix E (computational details) & Removed. It listed
\texttt{GraphRicciCurvature} among the software used, but no released
pipeline script calls that library (E1). \\
\bottomrule
\end{tabularx}
\caption{Status of the tables, figures and hypotheses of version~2.}
\label{tab:status}
\end{table}

\end{document}